%% file: iclr2023_conference.tex
\documentclass{article} 
\usepackage{iclr2023_conference,times}

\input{math_commands.tex}

\usepackage{hyperref, booktabs, bbding, multirow, graphicx}
\usepackage{url}
\usepackage{subcaption}
\hypersetup{
    colorlinks=true,
    linkcolor=blue,
    filecolor=magenta,      
    urlcolor=blue,
    pdfpagemode=FullScreen,
    }

\title{Compound Tokens: Channel Fusion for Vision-Language Representation Learning}

\author{Maxwell Mbabilla Aladago \thanks{Work done while interning at Google.} \\
Department of Computer Science\\
Dartmouth College\\
Hanover, NH 03755, USA \\
\texttt{\{maxwell.m.aladago.gr\}@dartmouth.edu} \\
\And
AJ Piergiovanni\\
Google Research \\
\texttt{\{ajpierji\}@google.com} \\
}

\iclrfinalcopy 
\begin{document}

\maketitle

\begin{abstract}
We present an effective method for fusing visual-and-language representations for several question answering tasks including visual question answering and visual entailment.
In contrast to prior works that concatenate unimodal representations or use only cross-attention, we compose multimodal representations via channel fusion. By fusing on the channels, the model is able to more effectively align the tokens compared to standard methods. These multimodal representations, which we call compound tokens are generated with cross-attention transformer layers. First, vision tokens are used as queries to retrieve compatible text tokens through cross-attention. We then chain the vision tokens and the queried text tokens along the channel dimension. We call the resulting representations compound tokens. A second group of compound tokens are generated using an analogous process where the text tokens serve as queries to the cross-attention layer. We concatenate all the compound tokens for further processing with multimodal encoder.  We demonstrate the effectiveness of compound tokens using an encoder-decoder vision-language model trained end-to-end in the open-vocabulary setting. Compound Tokens achieve highly competitive performance across a range of question answering tasks including GQA, VQA2.0, and  SNLI-VE.
\end{abstract}

\section{Introduction}
\label{introduction}
Multimodal learning will continue to play an increasingly fundamental role as we build increasingly more general purpose artificial agents. Tasks that seek information about visual inputs based on text queries such as visual question answering (VQA)~\citep{vqa2017, hudson2018gqa} have emerged as effective frameworks for multimodal learning as they require a thorough understanding of both visual and textual information. For example, to correctly answer the question ``what type of drink is to the right of the soda bottle?", a model must be able to distinguish a soda bottle from other bottles, left from right, understand language, and recognize the drink in question. Thus, effectively mixing or fusing the joint representations is critical for these tasks that have enjoyed so much progress in recent years~\citep{albef, blip,vlmo2021, coca2022, simvlr2022, flamingo2022, Wang2022}. 

\begin{figure}[t]
  \centering
   \begin{subfigure}[b]{0.32\textwidth}
         \centering
         \includegraphics[width=\textwidth]{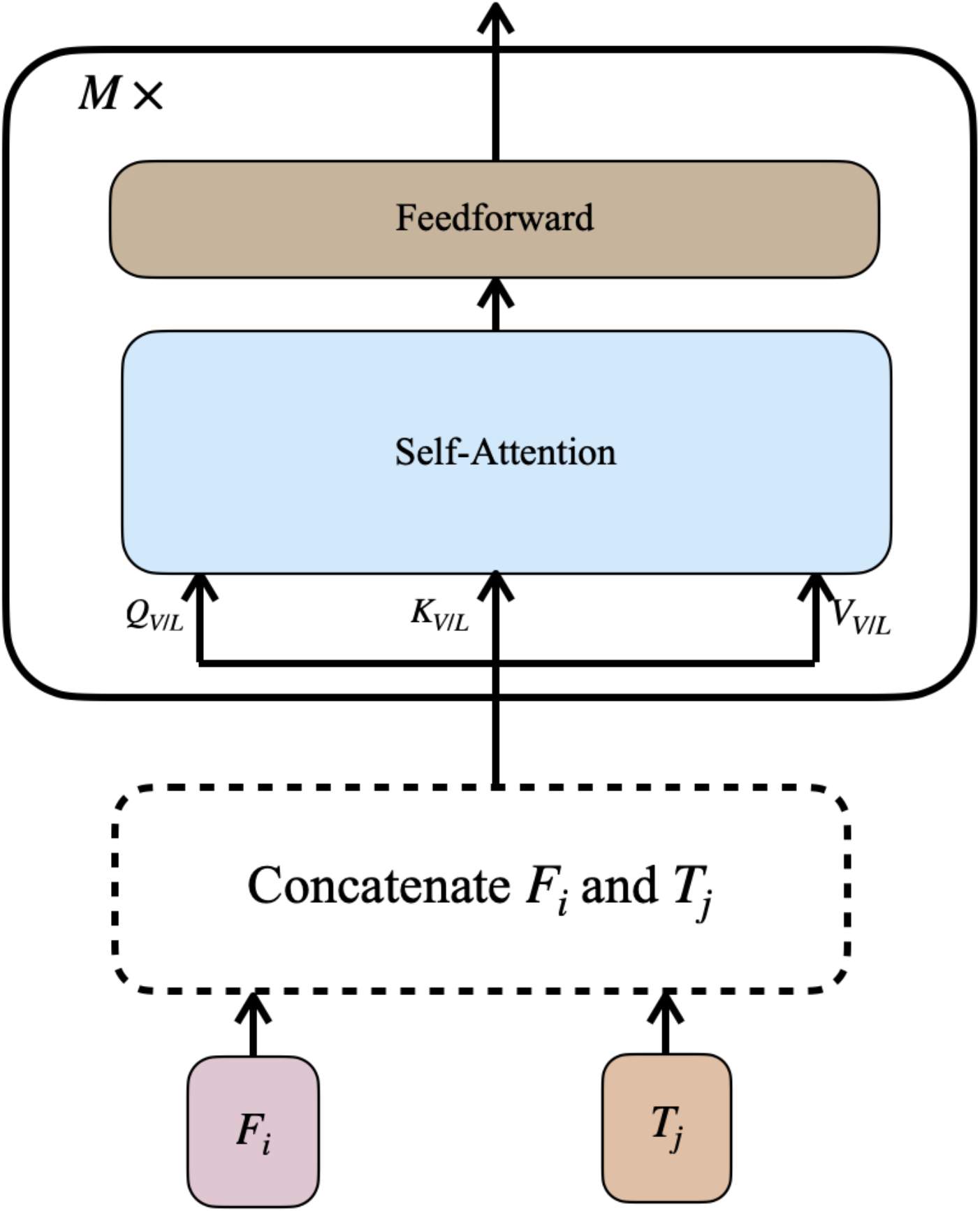}
         \caption{Merged Attention}
         \label{fig:merged-attention}
     \end{subfigure}
     \hfill
     \begin{subfigure}[b]{0.31\textwidth}
         \centering
         \includegraphics[width=\textwidth]{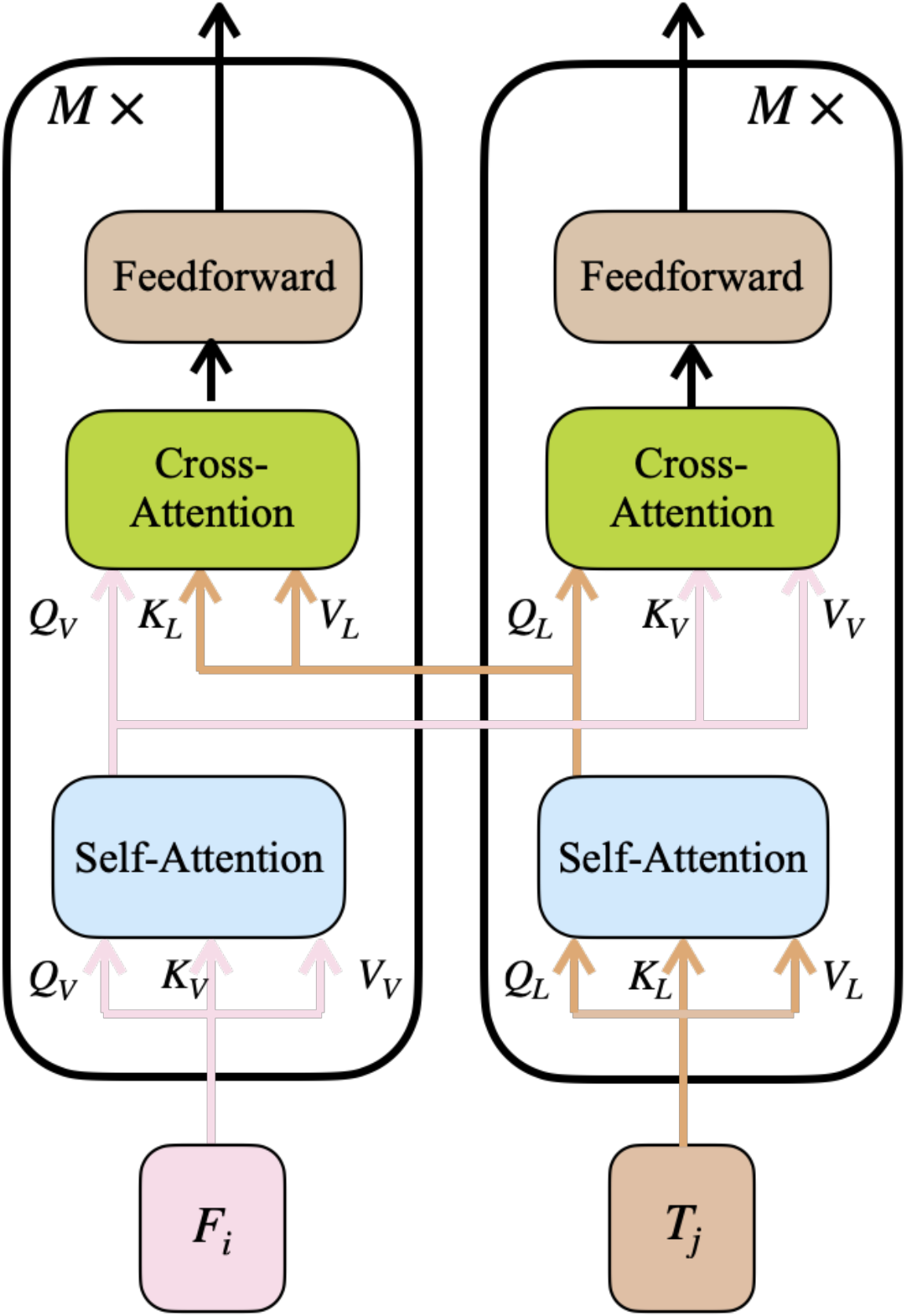}
         \caption{Co-Attention}
         \label{fig:co-attention}
     \end{subfigure}
     \hfill
     \begin{subfigure}[b]{0.32\textwidth}
         \centering
         \includegraphics[width=\textwidth]{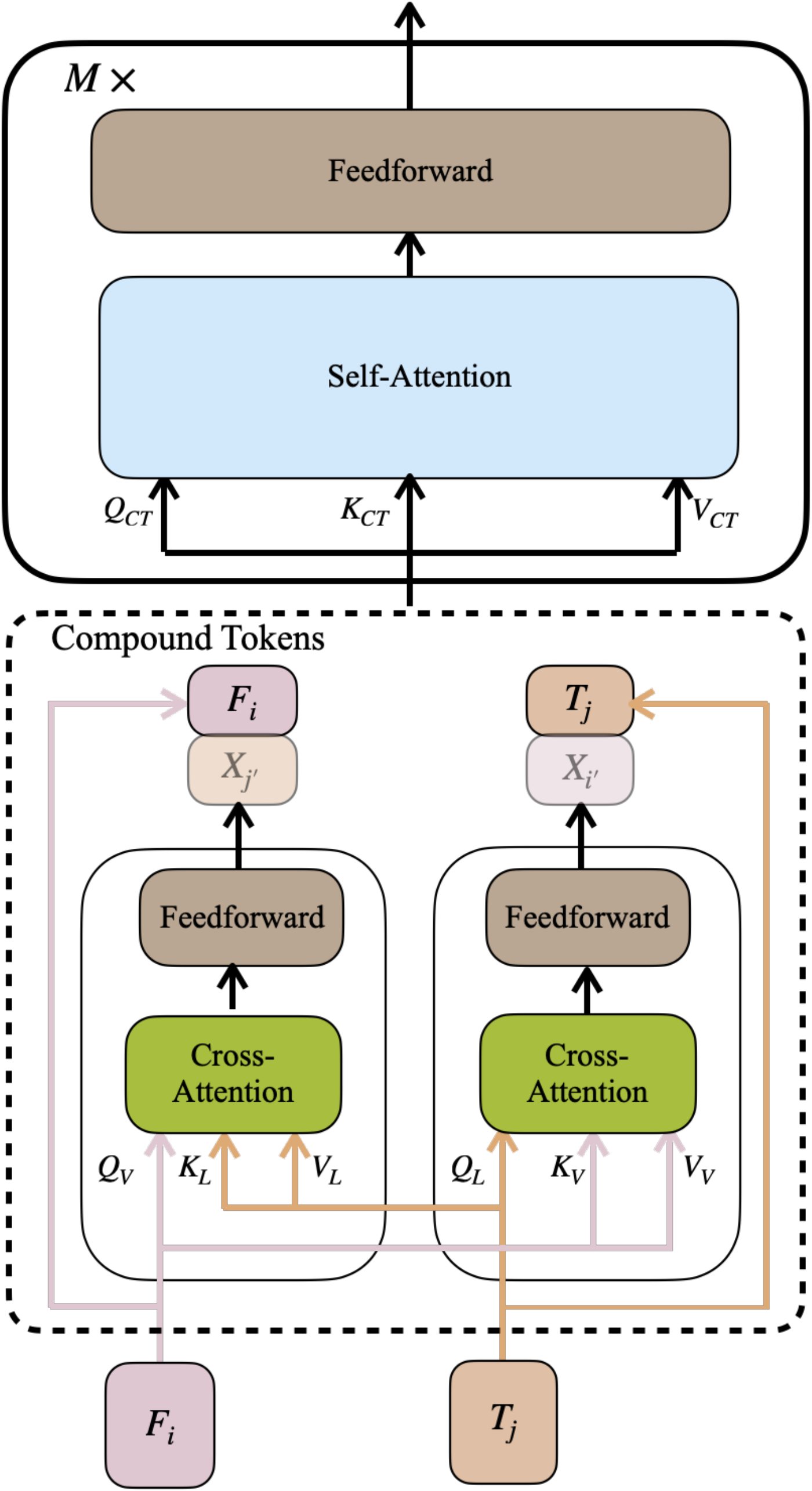}
         \caption{Compound Tokens}
         \label{fig:compound-tokens}
     \end{subfigure}
        \caption{\textbf{Different multimodal fusion methods}: Illustrations of two types of fusions methods in previous works: (a) co-attention, and (b) merged attention from the perspective of one visual token $F_i$, and one text token $T_j$. Our proposed compound tokens fusion method is illustrated in (c). Note that we use only one cross-attention layer for each modality compared to co-attention which uses both cross-attention and self-attention in all blocks. We concatenate the input query to the cross-attention module with the cross-attention output along the channel dimension. $Q$, $K$, and $V$ denote the input query, keys and values respectively to the attention module. $X$ represents the cross-attention layer's output. Finally, the subscripts $V$, $L$, and $CT$ respectively identify an input as visual features, text features or compound tokens, e.g., $Q_V$ indicates an input query that is composed of visual tokens. }
        \label{fig:fusion-methods}
\end{figure}

One common strategy for fusing multimodal representations is to simply concatenate the vision and text tokens together, and feed them into a multimodal transformer encoder with self-attention layers. This approach, which we use as our main comparison reference is christened \textit{merged attention}, and has been employed extensively in several vision-language models~\citep{zhou2019vlp, hendricks-etal-2021-decoupling, dou2022meter, answerme}. A second multimodal fusion method called \textit{co-attention} (shown in Figure~\ref{fig:fusion-methods}), feeds the text and visual tokens separately into independent transformer encoders, and leverages cross-attention to communicate information between the two modalities~\citep{tan2019lxmert, Bugliarello2021, hendricks-etal-2021-decoupling, albef, dou2022meter}.


While merge attention based models may struggle to align complementary tokens across different modalities effectively, co-attention based models forfeit the benefits of global self-attention across all tokens. Interestingly, \cite{dou2022meter} observed a performance boost from co-attention compared to merged attention, suggesting a beneficial effect from the cross-attention mechanism. The downside of co-attention, however, is that it is parameter inefficient compared to merged attention as it requires separate sets of parameters for vision and language features. Our work unifies merge attention and co-attention in an efficient pipeline that produces more powerful multimodal representations than either method for several multimodal tasks. 



Compound Tokens fusion aligns multimodal tokens using cross-attention without losing the advantages of global self-attention over all vision and text tokens. We use the tokens from one modality to query the other modality, and concatenate the output with the query tokens on the channels. An analogous process is repeated where we switch the roles of the two modalities. The resulting sets of compound tokens are concatenated and fed into a multimodal transformer encoder. Different from merged attention, we concatenate the vision and text tokens along the channel dimension. Unlike co-attention that uses both functions in every block, we use only two cross-attention functions at the beginning to facilitate channel concatenation. 

Combining the query features and the cross-attention outputs on the channels  (illustrated in Figure~\ref{fig:compound-tokens}) does not increase the token length, thus eliminating any additional computational or memory overheads in the multimodal transformer, and decoder modules. To further ensure that our method is efficient, we first embed each modality into half of their original feature dimension before compounding them. We show in Table~\ref{tab:different-combination-methods} that other ways of mixing the input queries and the cross-attention outputs, such as weighting or element-wise product, are less effective compared to channel concatenation.

\begin{figure}[t]
    \centering
    \includegraphics[width=\textwidth]{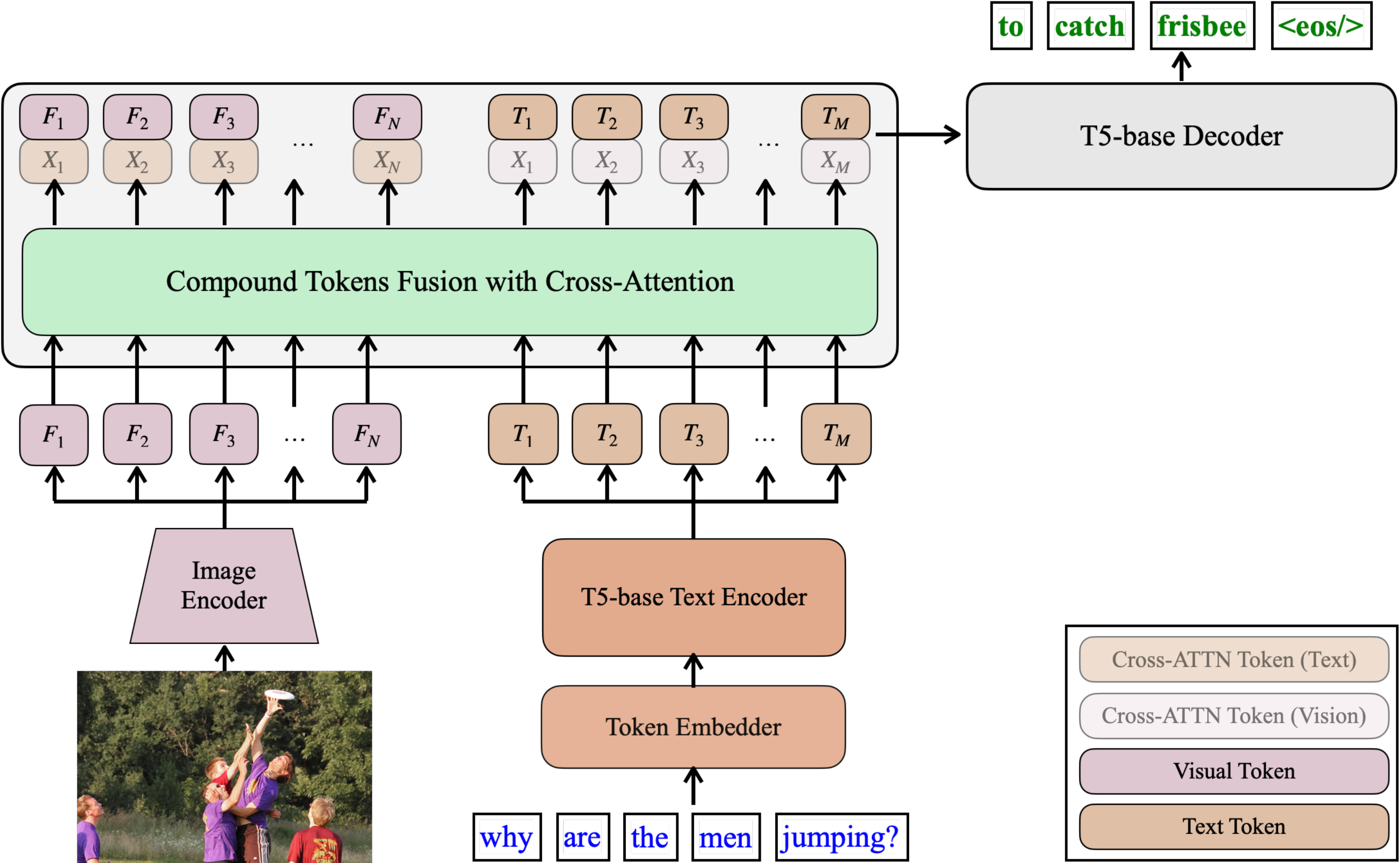}
    \caption{\textbf{Model Architecture:} Compound Tokens Fusion is illustrated in Figure~\ref{fig:compound-tokens}. ResNet-50~\citep{He2015} and T5-base~\citep{2020t5} are used for the image and text encoders respectively. }
    \label{fig:model-architecture}
\end{figure}

We evaluate compound tokens through extensive experiments in the challenging open-vocabulary setting via exact matching. In this setting, the generated responses must match exactly with the ground truth answers to be considered correct. This is notably more difficult than making predictions from a small predefined set of responses as in encoder only models. We adopt a generation pipeline following works such as \cite{pmlr-v139-cho21a}; \cite{simvlr2022}; and~\cite{ answerme} that demonstrated the flexibility of that approach and its relevance to practical scenarios. However, as observed in~\cite{dou2022meter}, this setting is less suitable for small models like ours. Accordingly, we provide separate results for VQA~\citep{vqa2017} using an encoder only model. Even in the encoder only setting, our pretraining setup still uses an encoder-decoder architecture as in~\cite{answerme} and illustrated in Figure~\ref{fig:model-architecture}.

Compound Tokens fusion outperforms both merged attention and co-attention on GQA~\citep{hudson2018gqa}, SNLI-VE~\citep{snlive}, and VQA~\citep{vqa2017} with and without vision-language pretraining. Our proposed fusion method obtained $82.87\%$ on SNLI-VE beating METER~\citep{dou2022meter} by $2.26\%$. Additionally, Compound Tokens recorded $82.43\%$ on GQA significantly outperforming CFR~\citep{Nguyen2022Coarse} by $8.83\%$. Our VQA score of $70.62\%$ on the VQA dataset is competitive among existing models.




To summarize, our work contributes a novel multimodal fusion method for vision-language tasks that enjoys the benefits of both cross-attention and self-attention without substantial additional computational overhead. We show the superiority of the proposed method over other fusion methods across several question answering tasks.


\section{Related Works}
\label{related-works}

Similar to the remarkable impact models such as T5~\citep{2020t5}, BERT~\citep{devlin-etal-2019-bert}, and GPT-3~\citep{gpt-3} have had on natural language process by pretraining on large amounts of text data, multimodal models like VilBERT~\citep{VilBertNEURIPS2019}, BEiT-3~\cite{Wang2022},  SimVLM~\citep{simvlr2022}, Flamingo~\citep{flamingo2022}, and PaLI~\citep{pali} have increasingly shown significant advantages from pretraining on large scale and diverse multimodal data. Unsurprisingly, vision-and-language tasks including visual dialog~\citep{visdial2017, visdial2018, visdial2022}, visual reasoning~\citep{reasoning2017, reasoning2019}, entailment~\citep{snlive, chen2020uniter}, visual question answering~\citep{Stanislaw-vqa2015, vqa2017, Huaizu-vqa2020, simvlr2022}, caption generation~\citep{Anderson2017up-down, changpinyo2021cc12m}, and cross-modality retrieval~\citep{Junhua-retrieval2016, kamath-retrieval2021} have all made great strides in recent years.

Important architectural innovations in vision-and-language models have been instrumental in accelerating these impressive scaling capabilities. One such innovation is the switch from expensive object detectors such as Faster-RCNN~\citep{fast-rcnn} in earlier models~\citep{tan2019lxmert, VilBertNEURIPS2019, li2019visualbert, li2020oscar, Zhang2021CVPR} to simpler modules such as ResNet~\citep{He2015} or Vision Transformer~\citep{dosovitskiy2021an} for encoding visual features. Removing the object detectors reduced the need to train on clean human-annotated datasets such as Visual Genome~\citep{krishnavisualgenome}, thus paving the way to more impactfully use copious amounts of weakly-supervised image-text datasets from the internet.


Pretraining vision-and-language models with different cross-modal objectives has been another major axis of exploration in recent works. Contrastive learning~\citep{li-etal-2021-unimo, coca2022}, image captioning~\citep{Anderson2017up-down},  image-text matching~\citep{Lee2018, VilBertNEURIPS2019}, prefix language modeling~\citep{simvlr2022}, word-patch alignment~\citep{Kim2021}, etc., are some of the variety of losses proposed recently. Other works combine several losses during pretraining~\citep{blip, dou2022meter}, while still more methods unify several question answering tasks into a multi-task framework~\citep{multitask, Lu-2020-CVPR, answerme}.

While a majority of works on vision-and-language representation learning have concentrated on improving feature extraction of the distinct modalities (e.g., using an object detector versus a convolutional neural network or a Transformer for vision feature extraction) or devising novel objective functions, efforts on improving the fusion of the multimodal representations have garnered relatively little attention. Most researchers simply adopt concatenation as described in merged attention~\citep{dou2022meter} for fusion~\citep{zhou2019vlp, answerme, simvlr2022}. These methods differ sometimes on whether merging is done early in the model or at a deeper stage after processing each modality with large independent backbones. Co-attention is another popular method for mixing the multimodal features~\citep{tan2019lxmert, albef, dou2022meter}. In co-attention, the vision and text features are modeled independently in separate transformer encoders, with a cross-attention mechanism serving as the bridge between the two modalities. 


This work focuses on improving the multimodal fusion of the different representations for question answering tasks. As a result, we do not dwell heavily on the type of backbone encoders, the pretraining style nor the loss functions used. We propose a novel fusion of multimodal tokens through channel concatenation to encourage better alignment between the tokens, while preserving the capability to use self-attention over the joint representations. We use (1) a vision-to-text cross-attention, and (2) a text-to-vision cross-attention to retrieve more aligned representations. These aligned representations are concatenated with the query tokens on the feature dimension to form what we call compound tokens. The vision-to-text and text-to-vision compound tokens are merged and fed into a transformer encoder. See Figure~\ref{fig:fusion-methods} for illustrative comparisons between merged attention, co-attention and compound tokens fusion. 

\section{Compound Tokens}
\label{main-method}
\subsection{Background}
We now provide a high level background on relevant functions necessary for the understanding of our method. We ignore layer normalization and multi-layer perceptrons in attention blocks in this overview for simplicity. For the same reason, we do not discuss residual connections between layers.

\textbf{Attention: } Given a set of query vectors $\mathbf{Q}\in\mathbb{R}^{N \times d}$ and a set of key vectors $\mathbf{K} \in \mathbb{R}^{M \times d}$, an attention layer 
gathers information from context vectors $\mathbf{V} \in \mathbb{R}^{M \times c}$ proportional to the normalized scores between the elements of $\mathbf{Q}$ and $\mathbf{K}$. Specifically, for softmax dot-product attention~\citep{vaswani2017}, the scalar output $z_{i, \ell}$, of an attention layer for query vector $q_i \in \mathbf{Q}$ and key vector $k_j \in \mathbf{K} $, is the weighted sum of the elements of $\mathbf{V}$,
\begin{align}
    a_{i,j} &= \frac{q_i^Tk_j}{\sqrt{d}} & \alpha_{i,j} &= \frac{\exp({a_{i, j})}}{\sum_\ell\exp({a_{i,\ell})}} & z_{i, \ell} &= \sum_j\alpha_{i,j}\mathbf{V}_{j, \ell}  \ .
\end{align}
An attention mechanism is called self-attention when the query vectors are also members of the context vectors, i.e., $q_i \in \mathbf{K} \; \forall i$. It is known as cross-attention otherwise.

\textbf{Multimodal Fusion:} Token concatenation followed by self-attention is one of the most adopted approaches for cross-modal learning in recent vision-language architectures~\citep{li2019visualbert, answerme, simvlr2022, pali}. Formally, given a sequence of $N$ image tokens, $\mathcal{I} \in \mathbb{R}^{N \times d}$, and $M$ text tokens, $\mathcal{T} \in \mathbb{R}^{M \times d}$, most methods concatenate $\mathcal{I}$ and $\mathcal{T}$ into a single representation $\mathcal{O} \in \mathbb{R}^{(N + M) \times d}$ which is then fed into a multimodal transformer for further modeling. The target outputs are produced using either a linear layer or a decoder. Besides concatenation, other methods such as~\citep{tan2019lxmert, albef, blip} use multimodal transformers composed of both self-attention and cross-attention in every block. 



\subsection{Proposed Fusion Method}\label{fusion-method}
Our method, illustrated in Figures~\ref{fig:compound-tokens} \&~\ref{fig:model-architecture}, draws from both co-attention and merged-attention. Compound Tokens fusion first projects the visual and language tokens into half of the embedding space so that the total number of features is maintained after channel concatenation: $\widetilde{\mathcal{I}} \in \mathbb{R}^{N \times \frac{d}{2}}$; $\widetilde{\mathcal{T}} \in \mathbb{R}^{M \times \frac{d}{2}}$ for the image and text tokens respectively. Next, we employ only two cross-attention layers (unlike co-attention~\citep{dou2022meter} that uses cross-attention and self-attention in every block) to create visual and language compound tokens 
\begin{align}
    \widehat{\mathcal{I}} &= \mathcal{A}\left(\mathcal{\widetilde{I}}, \mathcal{\widetilde{T}}, \mathcal{\widetilde{T}}\right) & \in \mathbb{R}^{N \times \frac{d}{2}} & &
   \widehat{\mathcal{T}} &= \mathcal{A}\left(\mathcal{\widetilde{T}}, \mathcal{\widetilde{I}}, \mathcal{\widetilde{I}}\right) & \in \mathbb{R}^{M \times \frac{d}{2}} \\
   \mathcal{I}_{cmpd} &= \text{C-Concat}\left(\mathcal{\widetilde{I}}, \mathcal{\widehat{I}} \right) & \in \mathbb{R}^{N \times d}  & &
   \mathcal{T}_{cmpd} &= \text{C-Concat}\left(\mathcal{\widetilde{T}}, \mathcal{\widehat{T}} \right) & \in \mathbb{R}^{M \times d} \ ,
\end{align}
where $\mathcal{A}(q, k, v)$ is the cross-attention function with $q$, $k$, and $v$ as queries, keys, and values respectively.  $\text{C-Concat}(u, \upsilon)$ concatenates tensors $u$ and $\upsilon$ along the feature dimension. We combine vision-to-text compound tokens $\mathcal{I}_{cmpd}$, and text-to-vision compound tokens $\mathcal{T}_{cmpd}$, into a set of output compound tokens as in merged attention architectures
\begin{align}
     \mathcal{O}_{cmpd}  &= \text{Concat}\left(\mathcal{I}_{cmpd}, \mathcal{T}_{cmpd}\right) \in \mathbb{R}^{(N + M) \times d} \ . 
\end{align}
Following previous methods, we feed $\mathcal{O}_{cmpd}$ into a self-attention multimodal encoder before generating the target outputs with an auto-regressive decoder. We also show results in Figure~\ref{fig:merged-attention-cmpt-tkns-no-vlp} and Table~\ref{tab:merged-attention-cmpt-tkns-vlp} where we do not use a multimodal encoder: $\mathcal{O}_{cmpd}$ is passed directly into the decoder to produce the outputs.



\section{Experimental Setup}
\label{experimental-setup}

\subsection{Model}\label{model}
We use ResNet-50~\citep{He2015} as our image encoder and T5-base~\citep{2020t5} as our text encoder. The output of the image and text encoders are provided to our novel fusion method described in Section~\ref{fusion-method}. A T5-base decoder consumes the output of the fusion module and generates free form text for all question answering tasks. The image encoder is pretrained on ImageNet-1k~\citep{imagenet} while the text encoder and decoder use pretrained T5 weights. 

\subsection{Datasets}\label{datasets}

\textbf{SNLI-VE}~\citep{snlive} is  a dataset of approximately 500,000 image-text pairs used for visual entailment
(VE). Given an image and a proposed statement, the task for this dataset requires determining whether the statement is neutral, entails, or contradicts the image. 

\textbf{Visual Question Answering (VQA2.0)}~\citep{vqa2017} is a widely used benchmark for many question-answering models and contains 400,000 image-text pairs spanning 3,130 output categories. Each image-question pair is associated with 10 answers.


\textbf{GQA}~\citep{hudson2018gqa} is a vision question answering dataset of complex compositional questions comprising scene-object relations formed from Visual Genome~\citep{krishnavisualgenome} with approximately 22 million question-answer pairs and 113 thousand images.



We emphasize that for all tasks, our model must generate a correct answer in an open-vocabulary setting of about 32,000 words irrespective of the number of categories in the task. A generated response is counted as correct if and only if it matches exactly with the ground-truth answer. We use the VQA metric\footnote{https://visualqa.org/evaluation.html} for VQA2.0 and simple accuracy for GQA and SNLI-VE as evaluation metrics.


In addition to the downstream datasets, we also use CC3M\footnote{The version of the dataset we used has  about 2 million samples}~\citep{conceptualcaptions} and COCO Captions~\citep{mscoco} for pretraining. The pretraining setup uses a mixture of these datasets across four objectives:  (1) \textbf{image-text matching} where the model predicts whether an image-text pair is a match or not, (2) \textbf{captioning} where the model generates the full caption given an image, (3) \textbf{caption completion} where the model completes a masked caption, and (4) \textbf{masked-language modeling} as in BERT~\citep{devlin-etal-2019-bert}.




Unless otherwise stated, we pretrain our models  for 300k iterations using a batch size of 512 and perform an additional 100k iterations of finetuning at a batch size of 128 on the downstream tasks: SNLI-VE, VQA, and GQA. When pretraining, the image resolution is set to $224 \times 224$ which is increased to $384 \times 384$ during finetuning or when training from scratch without vision-language pretraining (VLP). The input text length is set to 32. The output text length is 32 during pretraining and reduced to 8 during finetuning. Table~\ref{tab:hyper-parameters} in the Appendix documents all our hyper-parameter settings including learning rates, weight-decay, etc. Generally, we use SNLI-VE and GQA for ablations as performance on those datasets in our setup is more stable than results on VQA. 


\section{Experimental Results }
\label{experimental-results}

\subsection{Why Channel Concatenation?}
To determine the best way of composing compound tokens, we examined a number of options with a prime objective to not increase the token length. To this end, we sampled four combination methods and compared them on SNLI-VE and GQA as the performances on these datasets in our setup are more stable compared to VQA. Given input queries $q$ and cross-attention layer's outputs $X$, we explored the following: (1) \textit{channel concatenation} where we concatenate $q$ and $X$ along the feature dimension as described in Section~\ref{fusion-method}. (2) \textit{weighting} uses the operation $Y = \alpha q + \beta X$ where  $\alpha$ and $\beta$ are learnable scalars initialized randomly, and $Y$ is the output. (3) In \textit{Element-wise product}, $Y = q \odot X$. (4) Finally, we tested a simple summation of the tensors, $Y = q + X$.  All these methods use approximately the same number of flops and parameters. The results in Table~\ref{tab:different-combination-methods} show channel concatenation is better than the other methods, hence our use of channel concatenation in the rest of the paper. 

\begin{table}[ht]
\caption{\textbf{Different Methods of Formulating Compound Tokens}: Channel concatenation obtains the highest accuracy on SNLI-VE and GQA.}
\label{tab:different-combination-methods}
\centering
\begin{tabular}{lc|ccc}
        \toprule
         Method & GFlops & SNLI-VE & GQA \\
         \midrule
         Channel Concatenation & 20.71 & \bf{80.85} & \bf{80.79} \\
          \midrule
          Weighting & 20.71 & 80.63 & 80.61\\
         \midrule
         Summation  & 20.71 & 80.75 & 80.35\\
         \midrule
          Element-wise Product &  20.71 & 80.81 & 78.31\\
        \bottomrule
    \end{tabular}
\end{table}

\begin{figure}[t]
    \centering
    \includegraphics[width=\textwidth]{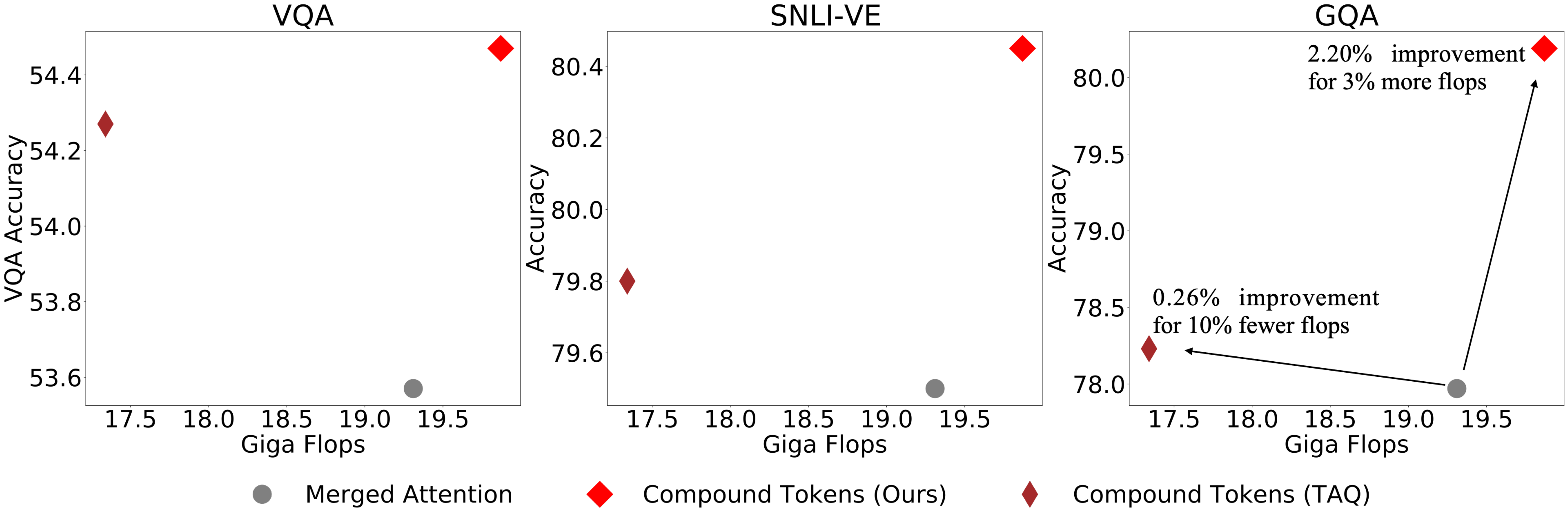}
    \caption{\textbf{Merged Attention versus Compound Tokens without Vision Language Pretraining: } With a relatively minimal amount of additional flops, Compound Tokens demonstrate a much improved performance over merged attention across all tasks. Compound Tokens (TAQ) is a more efficient version of our fusion method where we use only one cross-attention layer. For this method, we use the \textbf{T}ext tokens \textbf{A}s the input \textbf{Q}uery. We then compound the output with the text tokens along the channels as illustrated in Figure~\ref{fig:compound-tokens}. Note that even this more efficient compound tokens version outperforms merged attention, albeit with only marginal gains.}
    \label{fig:merged-attention-cmpt-tkns-no-vlp}
\end{figure}

\subsection{Comparison of Compound Tokens with Merged Attention}
We first compare merged attention and compound tokens fusion (our method) in Figure~\ref{fig:merged-attention-cmpt-tkns-no-vlp} without vision-language pretraining to establish some baseline results. We then incorporate vision-language pretraining and reassess the performance of each method in Table~\ref{tab:merged-attention-cmpt-tkns-vlp}. All three downstream tasks for each fusion method uses the same pretrained model.

For these baseline comparisons, the fusion modules do not use a multimodal encoder. Merged attention simply feeds a concatenation of the multimodal tokens to the decoder while compound tokens fusion passes the tokens to the decoder immediately after channel chaining.

The results in Figure~\ref{fig:merged-attention-cmpt-tkns-no-vlp} and Table~\ref{tab:merged-attention-cmpt-tkns-vlp} show clearly that compound tokens fusion is superior to merged attention with and without vision-language pretraining at a relatively small amount of additional compute. This performance boost suggests that the use of cross-attention to align the multimodal representations is positively impactful on performance for various question answering tasks. When vision-language pretraining is employed, Compound Tokens fusion outperforms merged attention by substantial margins on VQA ($+4.18\%)$ and GQA ($2.20\%$). The improvement on SNLI-VE is a relatively modest $0.24\%$. Our method enjoys similar improvement margins when training from scratch without vision-language pretraining. We include a more efficient version of our method (Compound Tokens (TAQ)) where we use only the text tokens as queries to these baseline comparisons. Even this reduced capacity version of our method outperforms merged attention across all tasks when training from scratch while using fewer flops. The ablation in Table~\ref{tab:abla-resolution} in the Appendix shows that the ranking here is consistent across different image resolutions. In Table~\ref{tab:image-encoder-ablation}, we show that Compound Tokens also beat merged attention when a Vision Transformer~\citep{dosovitskiy2021an} is used as the image-encoder instead of the ResNet-50 we used in these experiments. 



\begin{table}[t]
\caption{\textbf{Merged Attention versus Compound Tokens \textit{with} Vision Language Pretraining:} We repeat the experiments in Figure~\ref{fig:merged-attention-cmpt-tkns-no-vlp}, but include vision-language pretraining on a mixture of CC3M and COCO captions. While pretraining generally increases performance across all methods, compound tokens continue to surpass merged attention, further underscoring the superiority of our method.}
\label{tab:merged-attention-cmpt-tkns-vlp}
\centering
\begin{tabular}{lc|ccc}
        \toprule
         Fusion Method & GFlops & VQA & SNLI-VE & GQA \\
         \midrule
          Merged Attention  & 19.31 & 53.33 & 81.25 & 78.25\\
          Compound Tokens (Ours) & 19.87 & \bf{57.51} & \bf{81.49} & \bf{80.45}\\
          Compound Tokens (TAQ, Ours) & 17.34 & 53.23 & 81.21 & 77.74\\
        \bottomrule
    \end{tabular}
\end{table}

\begin{table}[ht]
    \caption{\textbf{Comparisons with other Fusion models \textit{without} Vision-Language Pretraining}: We extend the models to include a multimodal encoder with 12 self-attention layers in merged attention to match the typical setting in previous works. Compound Tokens outperform merged attention and Co-Attention with fewer parameters than both methods and fewer flops than merged attention. Co-Attention and merged attention are from ~\cite{dou2022meter} while Co-Tokenization is from~\cite{co-tokenization}. The results here are our implementations of the these methods.  Params shows the number of parameters in the entire model (not just the fusion module); RES is the image resolution and $L$ is the total number of transformer blocks in the multimodal encoder: Compound Tokens uses two cross-attention blocks before the multimodal encoder.}
\label{tab:merged-attention-cmpt-tkns-no-vlp-12}
    \centering
    \begin{tabular}{lcccc|cc}
        \toprule
         Fusion Method & $L$ & Params ($\times 10^6$) & RES & GFlops & SNLI-VE & GQA\\
         \midrule
        Merged Attention & 12 & $332.94$ & $384 \times 384$ & 34.89  &  79.81 & 78.07\\
        Co-Attention & 12 & $361.26$ & $384 \times 384$ & 29.61 & 80.20 & 77.75\\
        Compound Tokens (Ours) & 10 & $325.82$ & $384 \times 384$ & 32.90  &  \bf{80.52}  &  \bf{78.21}\\
        \midrule
        Co-Tokenization & 12 & $392.14$ & $384 \times 384$ & 57.78 & \bf{80.79} & \bf{81.07}\\ 
        Compound Tokens (Ours) & 10 & $325.82$ & $384 \times 384$ & 32.90  &  \underline{80.52}  &  \underline{78.21}\\
        \bottomrule
    \end{tabular}
\end{table}


\subsection{Multimodal Transformer Encoder}
After establishing the superiority of compound tokens over merged attention when no multimodal encoder is used before the decoder, we expand the models to include a multimodal encoder with 12 self-attention blocks to match the setting in most previous vision-language models~\citep{albef, dou2022meter}. We also compare with two other fusion methods Co-Attention (illustrated in Figure~\ref{fig:co-attention}), and Co-Tokenization~\citep{co-tokenization}. Originally implemented for question answering tasks in videos, Co-Tokenization iteratively fuses visual features with text features using a TokenLearner~\citep{token-learner}. We use an adaptation of Co-Tokenization for images. The Co-Attention fusion module uses 6 blocks each for the vision and the text branches as in METER~\citep{dou2022meter} where each block has a self-attention, cross-attention and feedforward layers. Co-Tokenization uses 64 image tokens and 4 transformer blocks per each tokenization round. There are 3 tokenization rounds, constituting 12 self-attention blocks overall. A self-attention block in our implementation is made up of a self-attention function and a feedforward layer. The multimodal encoder for Compound Tokens fusion has 10 blocks to compensate for the two cross-attention blocks that it uses. 

The results of these experiments are shown in Table~\ref{tab:merged-attention-cmpt-tkns-no-vlp-12}. The models are trained for 300k iterations at a batch size of 128 on each downstream task without any vision-language pretraining (See Table~\ref{tab:merged-attention-cmpt-tkns-vlp-12} in the Appendix for results with pretraining across different resolutions). Compound Tokens fusion continues to outperform merged attention and co-attention in this setting as well, indicating the fusion mechanism remains competitive even when a multimodal encoder is used. However, it slightly underperforms the more expensive Co-Tokenization module when training from scratch. 



\subsection{An Encoder only Model for VQA}
The performance of our models on VQA in the encoder-decoder setup is significantly lower than reported results in previous works even for small models like ours. We note again that this suboptimal performance is not unique to compound tokens fusion; we observe similar low values for all the fusion methods we tested in our architectural setup (See Table~\ref{tab:encoder-decoder-vqa} in the Appendix for VQA results in the encoder-decoder architecture for all fusion methods.). We believe the low performance is an effect of the decoder not being able to learn the VQA vocabulary sufficiently. To address any problems introduced by the decoder, we use an encoder only model for the VQA task during finetuning by replacing the decoder in a pretrained model with a linear layer of size $3130$. The results in Table~\ref{tab:vqa-classification} show that the encoder only model significantly outperforms the encoder-decoder model.  We, thus, adopt that setup in our comparison with previous work discussed in the next section. The VQA metric is still used for evaluation in the encoder only model. 


\begin{table}[ht]
    \caption{\textbf{Encoder only versus Encoder-Decoder}: The Encoder only model outperforms the encoder-decoder model by a large margin. The models here use the same pretrained encoder-decoder model: the decoder is replaced with a linear classifier when transitioning to an encoder-only version. We finetune both models for 100k steps after pretraining for 300k steps.}
\label{tab:vqa-classification}
    \centering
    \begin{tabular}{lcc|c}
        \toprule
         Fusion Method & Architecture & GFlops & VQA Accuracy\\
         \midrule
        Compound Tokens & Encoder-Decoder & 35.50 &  58.14\\
        Compound Tokens & Encoder Only &  31.77 & 70.39\\
        \bottomrule
    \end{tabular}
\end{table}

\subsection{Comparison with Existing Approaches}
Finally, we compare our results with various competitive recent models such as METER~\citep{dou2022meter}, ALBEF~\citep{albef}, and CFR~\citep{Nguyen2022Coarse}. The models in Table~\ref{tab:comparison-with-sota} generally have approximately the same number of parameters, but differ significantly on the pretraining datasets, pretraining objectives, and backbone encoders. For example, while we use Conceptual Captions~\citep{conceptualcaptions} and COCO~\citep{mscoco} as our pretraining datasets, METER used Conceptual Captions, COCO, Visual Genome~\citep{krishnavisualgenome} and SBU Captions~\citep{sbu-captions}. ALBEF used all the datasets in METER in addition to Conceptual Captions 12M~\citep{changpinyo2021cc12m}.

The model we use for this comparison has 340 million parameters in total. We pretrain it for 500k iterations with a batch-size of 512 using an image resolution of $224 \times 224$ and further finetune for 200k iterations on each of the downstream tasks at resolution $384 \times 384$ with batch size 128. This model uses a multimodal encoder with 12 blocks. 

Except for SimVLM~\citep{simvlr2022} which has about 1.5 billion parameters and uses a significantly large pretraining data (a 1.8 billion private dataset), our model outperforms all other methods on SNLI-VE and GQA by large margins. We are confident that further pretraining and increasing image resolution will improve our already competitive result on the VQA dataset. Scaling up the model may also yield additional performance improvements.  

\begin{table}[!ht]
\caption{\textbf{Comparison with SOTA:}  Compound Tokens outperforms all other models on SNLI-VE and GQA in an open-vocabulary evaluation except SimVLM~\citep{simvlr2022} which used a privat dataset of 1.5B samples. For VQA, we present the results in the closed-vocabulary setting for fair comparisons with the other methods: our open-set evaluation is significantly worse than the closed-set evaluation model on this task. The best values among the models besides SimVLM are in \textbf{bold}. The second best values are \underline{underlined}. *The flops are based on our calculations. Our model is extremely more efficient than the rest partly because we use a short text sequence length of 32 and a ResNet-50 backbone that produces 49 visual tokens. }
\label{tab:comparison-with-sota}
    \centering
    \begin{tabular}{l|cc|ccc}
    \toprule
        Approach & Params & GFlops$^*$ & VQA & SNLI-VE & GQA\\
        \midrule
         SimVLM$_{Huge}$~\citep{simvlr2022} & 1.5B & 890& \textit{\textcolor{gray}{80.34}} & \textit{\textcolor{gray}{86.32}} &- \\
         VisualBERT~\citep{li2019visualbert} &  & & 66.70 & 75.69 & -\\ 
         UNITER~\citep{chen2020uniter} &  & & 73.82 & 79.39 & - \\
         LXMERT~\citep{tan2019lxmert}  &  & & 69.90  & - & 60.00 \\
         ALBEF~\citep{albef}  & 418M & 122 & 75.84 & \underline{80.91} & -\\ 
         METER~\citep{dou2022meter} &336M & 130 &\textbf{77.68} & 80.61 & - \\
         BLIP~\citep{blip} & 475M & 122 &\underline{77.54} & - & - \\
         12-in-1~\citep{Lu-2020-CVPR} & & & 71.30 & - & 60.50 \\
         VinVL~\citep{Zhang2021CVPR} &  & & 75.95 & - & 65.05 \\ 
         VL-T5~\citep{pmlr-v139-cho21a} &  & & 70.30 & - & 60.80 \\
          CFR~\citep{Nguyen2022Coarse} & & & 69.80 & -& \underline{73.60} \\
         Compound Tokens (Ours) & 340M & 36 & 70.62 & \bf{82.87} & \bf{82.43} \\
         \bottomrule
    \end{tabular}
\end{table}

\section{Conclusion}
\label{conclusion}
We introduce Compound Tokens, a new multimodal fusion method for vision-and-language representation learning.  Our method beat super competitive models such as ALBEF and METER on SNLI-VE by close to 2\%. Furthermore, Compound Tokens performance on GQA beats the next best model we are aware of by more than 8 percentage points. Finally, we demonstrated through numerous comparative experiments that our method is better than merged attention and co-attention across three popular question answering tasks. We consistently outperformed these standard methods with and without pretraining on image-text pairs, across different image resolutions and image encoding backbones. 


With this strong demonstration as an effective fusion method, we hope that Compound Tokens will inspire other methods of modeling multimodal representations beyond token concatenation. 



\vspace{2cm}

\bibliography{iclr2023_conference}
\bibliographystyle{iclr2023_conference}

\vspace{2cm}
\appendix
\section*{Appendix}
\section{Hyper-parameter settings}\label{hyper-parameters}
We provide full details of our hyper-parameter settings in this section. We use Adam~\citep{adam} to optimize all our models. The learning rate starts from zero and warms up linearly to the base rate after 8k iterations. Cosine annealing~\citep{loshchilov2017sgdr} with a cycle rate of 100k steps is then used to decay the rate to zero by the end of training. We use gradient clipping with a maximum norm of $1.0$ in all our experiments.

We do not use any data augmentation beyond resizing and normalization in all the ablation experiments and finetuning experiments. We apply random cropping and AutoAugment~\citep{autoagmentation} during pretraining of our main model. 

All our pretraining experiments use a batch size of 512 and image resolution $224 \times 224$. The batch size is divided equally among the four pretraining objectives: image captioning, caption completion, image-text matching, and masked language modeling. We also sample the same number of examples from CC3M and COCO in every iteration. The batch size and resolutions are set to 128, and $384 \times 384$ respectively whenever training from scratch or finetuning.

The datasets we used and our model are described in  Section~\ref{experimental-setup}. The rest of the hyper-parameters are listed in Table~\ref{tab:hyper-parameters}. 


\begin{table}[!ht]
\caption{\textbf{Hyper-parameter Settings}: We enumerate the hyper-parameters for our ablation experiments and main model. $L$ is the number of blocks in a multimodal encoder. Main Model is the model we used in Table~\ref{tab:comparison-with-sota} for comparison with existing works. }
\label{tab:hyper-parameters}
\centering
\begin{tabular}{l|c|ccccc}
        \toprule
         Experiment & Phase & $L$ & Iterations & LR  & Dropout & Weight Decay\\
         \cline{1-7}
          \multirow{5}{*}{Ablations} & Pretraining & 0 / 12 & 300k & $1.1e^{-4}$  & $1e^{-3}$ & $0.1$ \\
          \cline{2-7}
         & \multirow{2}{*}{Finetuning} & 0 & \multirow{2}{*}{100k} & $5e-5$ & $0$ &  \multirow{2}{*}{$1e^{-4}$}\\
         &  & 12 &  & $3.1e^{-3}$   & $1e^{-3}$ & \\
         \cline{2-7}
         & \multirow{2}{*}{Scratch} & 0 & \multirow{2}{*}{300k} & $7.5e^{-5}$ & \multirow{2}{*}{$1e^{-2}$} & \multirow{2}{*}{$1e^{-3}$} \\
         &  & 12 & & $3e^{-5}$ & &  \\
         \cline{1-7}
         \multirow{2}{*}{Main Model} & Pretraining & \multirow{2}{*}{12} & 500k & $1.1e^{-4}$  & \multirow{2}{*}{$1e^{-3}$} & $0.1$ \\
         & Finetuning &  & 200k &  $3e^{-5}$ &  & $1e^{-4}$ \\
        \bottomrule
    \end{tabular}
\end{table}
\section{Further Ablations}
\subsection{Image Resolution}
Increasing image resolution generally leads to better performance for various question answering tasks. As a consequence, most prior works use a larger resolution during finetuning compared to the pretraining resolution. For example, \cite{simvlr2022} pretrained at resolution $224 \times 224$ and finetuned at $480 \times 480$. In this work, we followed the setting in METER~\citep{dou2022meter} by pretraining and finetuning at resolutions $224 \times 224$  and $384\times 384$ respectively. We now investigate whether Compound Tokens also enjoy improved performance relative to merged attention at different resolutions in this section. 

The results of this ablation is shown in Table~\ref{tab:abla-resolution} for models without a multimodal encoder and in Table~\ref{tab:merged-attention-cmpt-tkns-vlp-12} for models with a multi-modal encoder. The models in Table~\ref{tab:abla-resolution} do not use any pretraining on paired image-text data while the models in Table~\ref{tab:merged-attention-cmpt-tkns-vlp-12} are pretrained on CC3M and COCO for 300k iterations. As in prior works,  increasing image resolution improves performance across all fusion methods and datasets. In all cases, Compound Tokens continue to outperform merge attention, further underlining the fact that our proposed method is more effective than traditional merge attention.

\begin{table}[!ht]
\caption{\textbf{Impact of Image Resolution without Vision-Language Pretraining}:  Increasing the resolution increases performance for both merged attention and compound tokens, with compound tokens continuing to outperform merged attention at both resolutions. \textbf{Bold} numbers shows the best results within each comparison setting.}
\label{tab:abla-resolution}
\centering
\begin{tabular}{lcc|ccc}
        \toprule
         Fusion Method & RES & GFlops & SNLI-VE & GQA \\
         \midrule
          Merged Attention & $224\times224$ & 9.94 & 78.70 & 75.62\\
         Compound Tokens  & $224\times224$ & 10.22  & \bf{79.59} & \bf{76.62}\\
         \midrule
         Merged Attention & $384\times384$ & 19.31 &  79.15 & 76.66\\
         Compound Tokens & $384\times384$ & 19.87 & \bf{80.44} & \bf{79.02}\\
        \bottomrule
    \end{tabular}
\end{table}

\begin{table}[ht]
    \caption{\textbf{Impact of Image Resolution with Vision-Language Pretraining}:  The results here our implementations of the various methods.  Params shows the number of parameters in the entire model (not just the fusion module). $L$ is the number of self-attention blocks overall in the multimodal encoder. RES is the image resolution during finetuning: we pretrain all models at resolution $224 \times 224$. Increasing resolution generally leads to better performance on both datasets. Compound Tokens outperform all other fusion methods across the two resolutions.}
\label{tab:merged-attention-cmpt-tkns-vlp-12}
    \centering
    \begin{tabular}{lcccc|cc}
        \toprule
         Fusion Method & $ L$ & Params ($\times 10^6$) & RES & GFlops & SNLI-VE & GQA\\
         \midrule
        Merged Attention & 12 & $332.94$ & $224 \times 224$ & 16.95  & 81.01 & 77.06\\
        Co-Attention & 12 & $361.26$ & $224 \times 224$ & 14.63 & 79.89 & 75.06\\
        Co-Tokenization & 12 & $391.27$ & $224 \times 224$ & 28.84 & 81.52 & 77.60\\ 
        Compound Tokens & 10 & $337.67$ & $224 \times 224$ &  16.55 & 80.44  &  77.55\\
        Compound Tokens & 12 & $339.97$ & $224 \times 224$ & 17.23  &  \bf{81.75} &  \bf{79.92}\\
        \midrule
         Merged Attention & 12 & $332.94$ & $384 \times 384$ & 34.89 & 81.78 & 78.13 \\
        Co-Attention & 12 & $361.26$ & $384 \times 384$ & 29.61 & 80.50 & 75.92\\
        Compound Tokens & 10 & 337.67. & $384 \times 384$ & 33.95 & 79.93 &  78.73\\
        Compound Tokens & 12 & $339.97$ & $384 \times 384$ & 35.50 & \bf{82.47} & \bf{79.55}\\
        \bottomrule
    \end{tabular}
\end{table}

\subsection{Type of Image Encoder}
The image encoder is an important component in vision-language models. While earlier models used object detectors such as Faster-RCNN, more recent models use either a CNN~\citep{lenet1998} or a  Vision Transformer (ViT)~\citep{vaswani2017} for image feature extraction. We used ResNet-50 for our main experiments and investigate the impact of using a transformer as the image encoder in this ablation. The results of using using a ViT as the image encoder is shown in Table~\ref{tab:image-encoder-ablation}. All models in that experiment use $224 \times 224$ as the image resolution.  A patch size of $16 \times 16$ was used for the ViT based models. The ViT based models perform slightly less than the comparable ResNet based models. As is the case with using ResNet-50, Compound Tokens fusion remains superior to merged attention here as well. 


\begin{table}[!ht]
\caption{\textbf{Impact of Image Encoder}: Both the ViT-base and ResNet-50 are pretrained on ImageNet but we do not use any additional image-language pretraining. All models are trained for 300k iterations. Compound Tokens obtains a higher accuracy than merged attention across all image encoders.}
\label{tab:image-encoder-ablation}
\centering
\begin{tabular}{l|l|ccc}
        \toprule
         Image Encoder & Fusion Method & SNLI-VE & GQA \\
         \midrule
          \multirow{2}{*}{ViT-base}   & Merged Attention & 77.44  &  74.02\\
          &  \bf{Compound Tokens} &  \bf{78.59} &  \bf{74.74}\\
            \midrule
            \multirow{2}{*}{ResNet-50}  & Merged Attention &  78.70 &  75.62 \\
          &  \bf{Compound Tokens} & \bf{79.59}  & \bf{76.62} \\
        \bottomrule
    \end{tabular}
\end{table}

\section{Encoder-Decoder VQA Model}
We observed a generally low performance on VQA in our encoder-decoder model across all fusion mechanisms. We believe this is the case because our decoder is unable to generalize well to the VQA vocabulary due to our limited pretraining dataset. Besides the low performance, we also noticed that this dataset is very sensitive to hyper-parameter changes such as learning rate and dropout in our models. Faced with these challenges, we removed VQA from our ablations as indicated in the main text and show the results in this section for completeness. 

\begin{table}[!ht]
\caption{\textbf{Encoder-Decoder VQA Accuracy}: The VQA results in our encoder-decoder setup are generally low for all fusion methods and very sensitive to learning rate and dropout changes.  }
\label{tab:encoder-decoder-vqa}
\centering
\begin{tabular}{l|cccc}
        \toprule
          Setup & Merged Attention & Co-Attention & Co-Tokenization & Compound Tokens  \\
          \midrule
          Scratch & 55.20 & 52.43 & 51.94 & 54.43 \\
          \midrule
         Pretrained & 47.92 & 45.04 & 53.29 & 55.83 \\
        \bottomrule
    \end{tabular}
\end{table}

\end{document}

%% file: math_commands.tex
\usepackage{amsmath,amsfonts,bm}

\def\eqref#1{equation~\ref{#1}}

\def\1{\bm{1}}

\DeclareMathAlphabet{\mathsfit}{\encodingdefault}{\sfdefault}{m}{sl}
\SetMathAlphabet{\mathsfit}{bold}{\encodingdefault}{\sfdefault}{bx}{n}

